\documentclass[lettersize,journal]{IEEEtran}
\usepackage{amsmath,amsfonts}
\usepackage{algorithmic}
\usepackage{amsmath,amssymb}  
\usepackage{hyperref}  
\usepackage{url}  
\usepackage{algorithm}  
\usepackage{adjustbox}
\usepackage{booktabs}
\usepackage{array}
\usepackage[caption=false,font=normalsize,labelfont=sf,textfont=sf]{subfig}
\usepackage{textcomp}
\usepackage{stfloats}
\usepackage{url}
\usepackage{verbatim}
\usepackage{graphicx}
\def\BibTeX{{\rm B\kern-.05em{\sc i\kern-.025em b}\kern-.08em
    T\kern-.1667em\lower.7ex\hbox{E}\kern-.125emX}}
\usepackage{balance}
\begin{document}
\title{HDCNet: A Hybrid Depth Completion Network for Grasping Transparent and Reflective Objects}

\author{Guanghu Xie, Mingxu Li, Songwei Wu, Yang Liu\textsuperscript{\dag}, Zongwu Xie, Baoshi Cao, Hong Liu
\thanks{\textsuperscript{\dag} Corresponding author:Yang Liu(liuyanghit@hit.edu.cn).}
\thanks{*This work was supported by the Natural Science Foundation of Heilongjiang Province for Excellent Young Scholars (Grant No. YQ2024E018) and the Youth Talent Support Program of the China (Grant No. 2022-JCJQ-QT-061).}
\thanks{All authors are with with the State Key Laboratory of Robotics and Systems, Harbin Institute of Technology, Harbin 150001, Heilongjiang, China
}
}


\maketitle

\begin{abstract}
Depth perception of transparent and reflective objects 
has long been a critical challenge in robotic manipulation. 
Conventional depth sensors often fail 
to provide reliable measurements on such surfaces, 
limiting the performance of robots in perception 
and grasping tasks. To address this issue, 
we propose a novel depth completion network, HDCNet, 
which integrates the complementary strengths of Transformer, 
CNN, and Mamba architectures. 
Specifically, the encoder is designed 
as a dual-branch Transformer–CNN framework 
to extract modality-specific features. 
At the shallow layers of the encoder, 
we introduce a lightweight multimodal fusion module 
to effectively integrate low-level features. 
At the network bottleneck, 
a Transformer–Mamba hybrid fusion module 
is developed to achieve deep integration 
of high-level semantic and global contextual information, 
significantly enhancing depth completion accuracy 
and robustness. Extensive evaluations 
on multiple public datasets demonstrate 
that HDCNet achieves state-of-the-art(SOTA) performance 
in depth completion tasks. Furthermore, 
robotic grasping experiments show 
that HDCNet substantially improves grasp success rates 
for transparent and reflective objects, achieving up to a 60\% increase.
\end{abstract}


\section{Introduction}


Three-dimensional (3D) depth information constitutes the foundation of intelligent perception systems for understanding the physical world, and its accuracy directly affects the performance of tasks such as robotic manipulation, autonomous navigation, 3D reconstruction, and human–robot interaction. With the increasing prevalence of transparent objects in industrial and daily applications, depth perception of such objects has attracted growing research interest\cite{9915452}\cite{9715114}\cite{9707827}. However, when objects exhibit transparent or highly reflective properties, conventional depth sensing techniques face significant challenges\cite{sun2024diffusion}. Transparent and reflective materials alter light propagation—through refraction, transmission, and specular reflection—thereby violating the Lambertian assumption underlying most traditional depth sensors.

Active depth sensors, including structured light, LiDAR, and time-of-flight (ToF) cameras, generally assume that emitted light is diffusely reflected back to the receiver. For transparent media, however, light is refracted or transmitted, leading to attenuation or signal loss, while reflective surfaces deflect illumination away from the receiver’s field of view, producing false or missing depth measurements\cite{dai2022domain}\cite{tian2023data}\cite{dong2022graspvdn}. Even with precise calibration, RGB-D systems often suffer from depth misalignment, double reflections, and discontinuities on non-Lambertian surfaces\cite{10377966}\cite{8166766}. These issues cause extensive holes in depth maps, resulting in degraded performance in downstream tasks such as 3D reconstruction, pose estimation, and robotic grasping\cite{huang2024distillgrasp}.

The fundamental reason for these deficiencies lies in the intricate coupling between optical and geometric phenomena \cite{10669217}. Refraction complicates the differentiation between object surfaces and their backgrounds. Specular reflections further obscure the true surface geometry. Moreover, multipath interference and ambient illumination can distort sensor responses and reduce measurement reliability \cite{10597633}. 

Due to the lack of geometric constraints, monocular RGB-based depth estimation remains inherently ill-posed \cite{li2024segment,shen2025gamba}. Consequently, modern research has progressively transitioned from hardware refinement to learning-based modeling, aiming to bridge the physical limitations of sensing through multimodal integration and data-driven reasoning.

\begin{figure}[htbp] 
    \centering
    \includegraphics[width=\columnwidth]{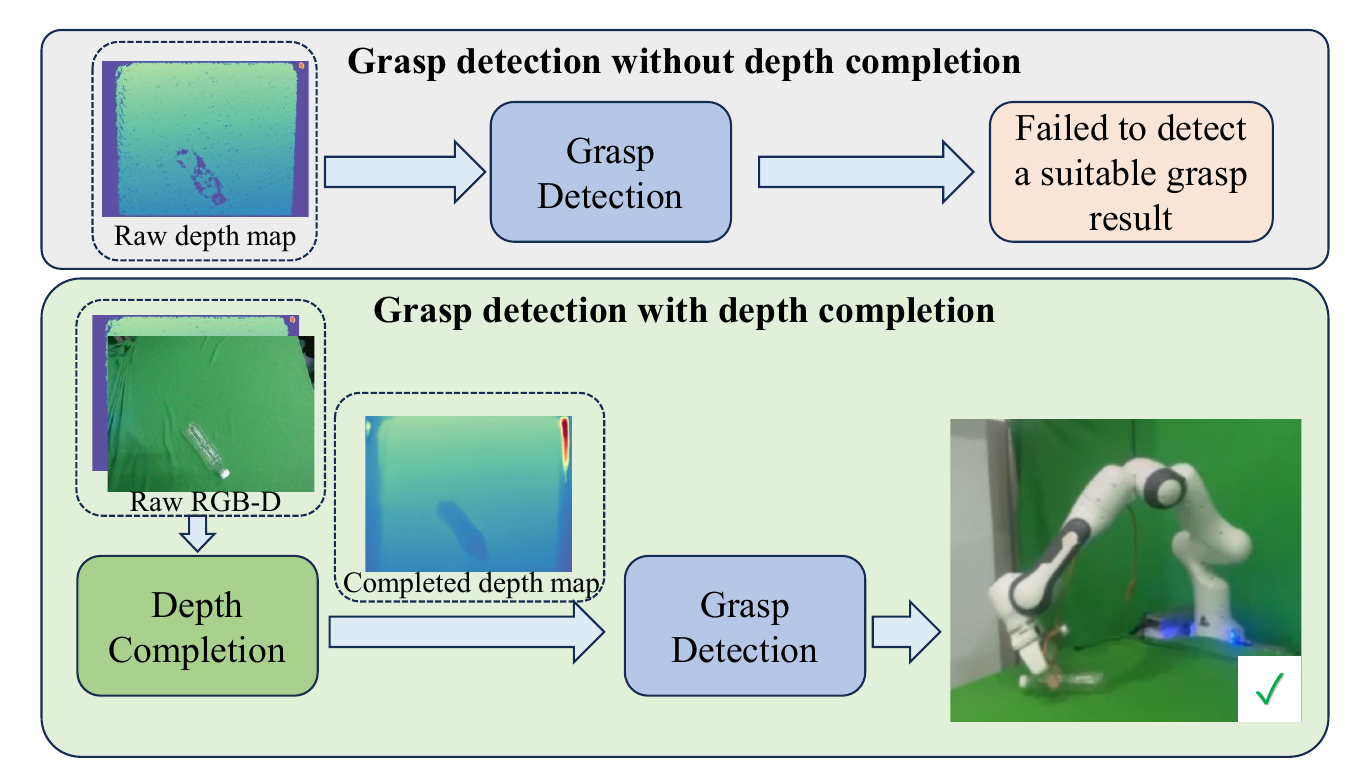} 
    \caption{Comparison of grasp detection for transparent and reflective objects with and without depth completion.}
    \label{intro_pic}
\end{figure}

Learning-based depth completion can effectively improve the grasp success rate of transparent and reflective objects, as shown in Fig.\ref{intro_pic}.
Learning-based depth completion and estimation frameworks have achieved remarkable progress in recent years. 
By leveraging neural representations, they can capture nonlinear relationships between appearance and geometry, enabling the recovery of dense and physically consistent depth structures even under challenging optical conditions. This paradigm shift replaces explicit optical modeling with learned priors that infer geometric continuity through high-dimensional feature correlation. Moreover, multimodal fusion and attention mechanisms enhance robustness in complex environments, allowing these systems to operate across diverse materials and lighting conditions\cite{liu2025gaa}\cite{dai2022domain}.

Despite these advances, major obstacles remain. Many models still struggle to generalize across varying materials, illumination, and viewpoints. The scarcity of high-quality transparent-object datasets constrains supervised learning, while purely synthetic data often fail to capture the full optical complexity of real-world transparency. In addition, architectural designs still rely on empirical heuristics rather than a principled understanding of cross-modal alignment. Thus, constructing more physically grounded feature representations, improving domain adaptation, and achieving efficient multimodal fusion remain core challenges for future research in transparent-object depth estimation.


To tackle the long-standing challenge of depth perception for transparent and reflective objects, this work proposes a hybrid depth completion framework that effectively integrates Transformer, CNN, and Mamba architectures. The main contributions of this paper are summarized as follows:

\begin{itemize}
    \item \textbf{Hybrid Depth Completion Network (HDCNet).}  
    We propose a novel depth completion network that unifies the complementary strengths of Transformer, CNN, and Mamba architectures, enabling both local structure preservation and global context modeling for complex visual scenes.
    
    \item \textbf{Hierarchical Multimodal Fusion Mechanism.}  
    We introduce a hierarchical fusion strategy that includes a lightweight shallow multimodal fusion module for early low-level feature alignment and a Transformer–Mamba hybrid deep fusion module at the bottleneck to achieve semantic-level and global contextual integration.
    
    \item \textbf{Comprehensive Evaluation and Real-world Validation.}  
    Extensive experiments on multiple public datasets demonstrate that HDCNet achieves state-of-the-art performance in depth completion tasks. Moreover, robotic grasping experiments validate its practical effectiveness, improving grasp success rates for transparent and reflective objects by up to 60\%.
\end{itemize}

\section{Realted Work}

\subsection{Depth Completion}
Early studies on transparent and reflective object reconstruction primarily relied on geometric and optical modeling approaches\cite{karras2019conference}, such as refraction equation solving and multiview stereo techniques\cite{chen2018estimating, cheng2018depth, fu2020depth, lee2021depth}. These methods require precise calibration, controlled lighting, and prior knowledge of refractive indices. However, under realistic conditions with layered transparency, interreflections, or environmental noise, their robustness degrades significantly\cite{qian20163d, wu2018full}. Moreover, traditional algorithms are computationally intensive and ill-suited for real-time robotic deployment.

With the increasing availability of high-resolution RGB-D sensors and GPU computing, learning-based methods have become dominant. They aim to learn the nonlinear mapping between sparse depth inputs and visual features through data-driven optimization\cite{chen2021transunet}. Such models are capable of leveraging global context, material appearance, and edge cues, thereby improving reconstruction fidelity where optical models fail.

Depth completion methods are broadly divided into two paradigms: direct regression and optimization-guided inference. Direct regression approaches employ CNNs or Transformers to predict dense depth maps directly from visual cues, while optimization-guided methods enforce geometric consistency and depth smoothness via loss regularization. GAN-based formulations\cite{9636382} generate photometrically consistent depth maps by learning structural priors, whereas skip-connected U-Net variants\cite{li2023fdct} enhance multiscale feature reuse for fine-grained details. Fang et al.\cite{fang2022transcg} further proposed an end-to-end U-shaped framework that jointly processes RGB and depth modalities, enabling complementary feature learning.
Yan et al.\cite{yan2024transparent} introduced TODPN, a transparent-object depth completion network that integrates geometry-aware constraints to refine grasp-oriented regions. 
This evolution from geometric modeling to neural inference represents a broader trend toward unified frameworks that couple photometric cues with learned physical priors.
\subsection{Hybrid Network Framework}
Convolutional Neural Networks (CNNs) have long dominated dense prediction tasks due to their strong local feature extraction capabilities \cite{eigen2014depth}. By exploiting spatially shared convolutional kernels, CNNs effectively capture fine-grained details such as edges, textures, and local geometric structures. However, their inherently limited receptive field and inductive bias toward locality restrict the ability to model long-range dependencies and complex geometric relationships across the entire scene. This limitation often results in fragmented depth predictions and discontinuities, especially under highly reflective or transparent conditions where global context is essential.

To address the aforementioned challenges, recent studies have increasingly explored the complementary strengths of Transformer and CNN architectures. For instance, in the field of medical imaging, TransUNet \cite{chen2021transunet} was the first model to integrate a U-shaped CNN with a Transformer for medical image segmentation, effectively combining local feature extraction with global contextual representation to improve segmentation accuracy and structural consistency.

The self-attention mechanism inherent in Transformer architectures dynamically models dependencies among all spatial positions, thereby facilitating global reasoning and enhancing contextual understanding. These attention-driven models exhibit superior capabilities in representing non-Lambertian light interactions and maintaining scene-level consistency. Particularly in transparent-object scenarios, Transformers leverage global correlations between appearance and geometry to effectively disambiguate refracted edges and specular highlights. Fan et al. \cite{11134066} further proposed a residual fusion design that combines CNN and Transformer encoders to strengthen cross-scale feature interactions and cooperative representation. This hybrid architecture achieves a better balance between fine-grained geometric details and holistic scene understanding while significantly reducing artifacts near reflective boundaries. Such integration fully exploits the local sensitivity of CNNs and the global modeling capability of Transformers, providing a more robust solution for depth estimation under complex illumination and transparent-object conditions.

Recent research trends point toward hybrid architectures that merge CNN’s locality with Transformer’s global attention \cite{khan2023survey}. These designs unify the strengths of convolutional inductive priors—translation equivariance and noise robustness—with the flexibility of attention-based global modeling. Such hybrid frameworks have gradually become the prevailing solution for transparent-object depth completion and reconstruction tasks \cite{chen2022mobile, wang2021evolving}. Beyond performance gains, this hybridization also supports better cross-domain generalization and robustness under real-world optical distortions, paving the way for unified vision–geometry reasoning networks that can simultaneously perceive fine local details and holistic scene structures.
\subsection{Multimodal Information Fusion}
In the perception and recognition of transparent and specular objects, relying on a single modality often fails to accurately capture the complex optical characteristics and geometric structures of the scene. Transparent materials cause light refraction and transmission, while specular surfaces produce strong reflections and highlights, resulting in significant measurement errors in traditional RGB- or depth-only systems. Specifically, RGB images contain rich texture, color, and semantic cues but are highly sensitive to illumination conditions, reflection strength, and background interference. Depth images, on the other hand, provide explicit geometric information yet frequently suffer from missing measurements, noise accumulation, and structural distortions in reflective or transparent regions\cite{jiang2025dcdu}. The inherent limitations of each modality make it difficult for single-modality systems to obtain reliable 3D representations under complex material and lighting conditions, thereby constraining their performance in tasks such as recognition, reconstruction, and pose estimation.

To address these challenges, multi-modal feature fusion has emerged as an effective strategy to integrate complementary information from different modalities. By introducing dual- or multi-branch encoder architectures, networks can simultaneously learn semantic cues from RGB data and geometric constraints from depth data. The fusion of these modalities enables a more comprehensive understanding of the scene, enhancing robustness against illumination variations, material diversity, and optical distortions. Methods such as \cite{qiu2019deeplidar,ma2018sparse} combine RGB images with sparse raw depth maps for depth completion or refinement. Compared with purely color-based approaches, these methods improve prediction accuracy; however, limited depth sampling can still result in low-quality outputs, such as blurred edges and local geometric inconsistencies.

Recent studies have increasingly focused on Transformer-based cross-modal fusion frameworks. 
These methods aim to integrate semantic and geometric information more effectively by enabling adaptive interactions between RGB and depth representations.
For instance, SwinDRNet~\cite{dai2022domain} employs two parallel Swin Transformer backbones to independently extract RGB and depth features. 
It then applies a cross-attention fusion module that adaptively aligns and integrates the two modalities, dynamically balancing modality contributions and suppressing redundant features.

\section{Method}

\begin{figure*}[htbp] 
    \centering
    \includegraphics[width=\textwidth]{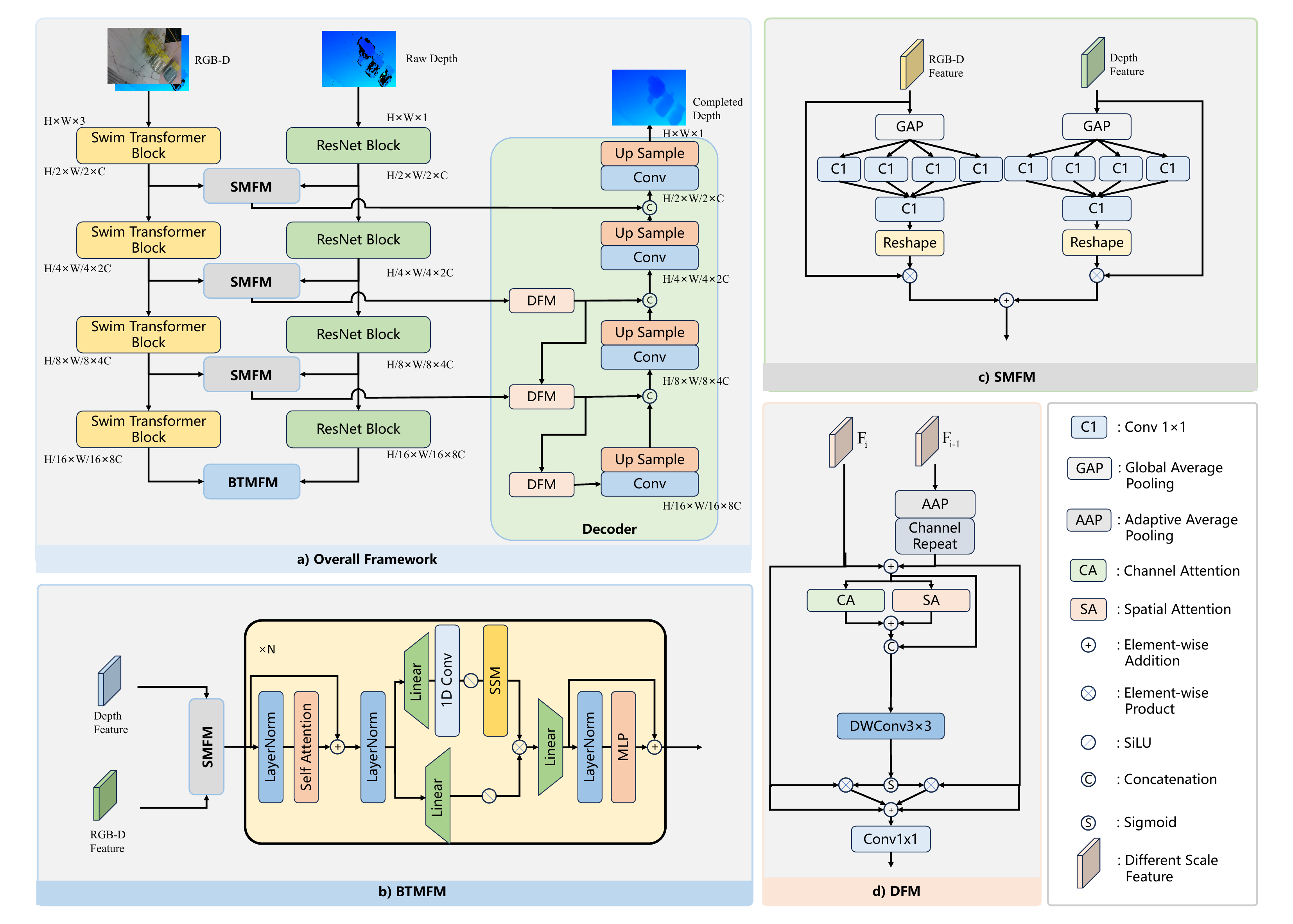} 
    \caption{HDCNet Architecture.Our method consists of a dual-branch encoder, a bottleneck fusion module, and a decoder. The Transformer-based backbone extracts RGB-D features, while the CNN-based backbone extracts depth features. The bottleneck fusion module performs multimodal fusion at the network bottleneck, and the decoder is composed of a multi-scale fusion module, convolutional layers, and upsampling operations. 
    }
    \label{fig1}
\end{figure*}

\subsection{Dual-branch Transformer-CNN encoder}
In recent years, hybrid architectures that combine Transformer 
and CNN branches have demonstrated remarkable effectiveness 
in depth completion tasks. As reported in \cite{fan2024tdcnet}, 
Transformer modules are particularly adept 
at capturing texture and semantic information from RGB-D images, 
while CNNs are better suited for modeling geometric structures 
in depth maps. This observation provides a strong rationale 
for complementary multimodal feature fusion. Motivated 
by this insight, we design a dual-branch encoder framework: 
the RGB branch incorporates a Swin Transformer
to leverage its strengths in modeling long-range dependencies 
and multi-scale contextual representations, 
thereby enhancing the extraction of semantic 
and texture features. In parallel, 
the depth branch employs a ResNet backbone 
to exploit its capability in capturing local structures 
and geometric cues. Such a design effectively 
harnesses the complementary advantages of Transformer and CNN, 
laying a solid foundation 
for subsequent cross-modal feature alignment and fusion.

\subsection{Shallow multimodal fusion module}
After being processed by the dual-branch encoders, 
multi-level feature representations 
are obtained at different stages. 
To effectively integrate information from different modalities into a unified fused representation, 
we propose a lightweight shallow multimodal fusion module(SMFM) specifically designed for shallow feature fusion. 

Formally, let $\mathbf{F}_{r} \in \mathbb{R}^{C \times H \times W}$ and $\mathbf{F}_{d} \in \mathbb{R}^{C \times H \times W}$ denote the feature maps extracted from two different modalities, where $C$, $H$, and $W$ represent the number of channels, height, and width, respectively. 
To highlight discriminative information and reduce redundancy, channel-wise feature enhancement is first applied to the feature maps using a multi-branch bottleneck structure.

Given a feature map $\mathbf{F}_{m}$, $m \in \{r,d\}$, the channel descriptor is first obtained via global average pooling:
\begin{equation}
    \mathbf{z}_{m} = \mathrm{GAP}\!\left(\mathbf{F}_{m}\right) \in \mathbb{R}^{C}.
\end{equation}

Then, four parallel linear transformations with learnable parameters $\mathbf{W}_1, \mathbf{W}_2, \mathbf{W}_3, \mathbf{W}_4$ are applied to $\mathbf{z}_{m}$ to compress the channel dimension in each branch. Their outputs are concatenated and then passed through a final linear transformation $\mathbf{W}_5$ to expand the channel dimension back to $C$:
\begin{equation}
    \mathbf{s}_{sq} = \mathrm{Concat}\Big( 
        \delta(\mathbf{W}_1 \mathbf{z}_{m}), \,
        \delta(\mathbf{W}_2 \mathbf{z}_{m}), \,
        \delta(\mathbf{W}_3 \mathbf{z}_{m}), \,
        \delta(\mathbf{W}_4 \mathbf{z}_{m})
    \Big),
\end{equation}

\begin{equation}
    \mathbf{s}_{ex} = \sigma(\mathbf{W}_5\mathbf{s}_{sq}),
\end{equation}

Where $\delta(\cdot)$ denotes the ReLU activation, $\sigma(\cdot)$ is the sigmoid function, and $\mathrm{Concat}(\cdot)$ concatenates the outputs of the four branches. 

The enhanced feature is subsequently computed by applying the channel attention vector to the original feature map:
\begin{equation}
    \tilde{\mathbf{F}}_{m} = \mathbf{s}_{ex} \odot \mathbf{F}_{m},
\end{equation}

Where $\odot$ denotes channel-wise multiplication. 

Finally, the enhanced features from the two modalities are fused through fine-grained element-wise addition:
\begin{equation}
    \mathbf{F}_{fused} = \tilde{\mathbf{F}}_{r} \oplus \tilde{\mathbf{F}}_{d},
\end{equation}

Where $\oplus$ represents element-wise summation. This multi-branch compression-expansion design enables complementary cues across modalities to be fully exploited, yielding a compact yet expressive fused representation that benefits downstream tasks.

\subsection{Bottleneck Transformer-Mamba fusion module}
At the bottleneck of the model, feature maps generally exhibit low spatial resolution. Performing feature fusion at this stage can effectively reduce computational cost while improving model performance, as demonstrated in previous works~\cite{yanhetero}.
To address this, we propose a novel multimodal fusion module that combines the strengths of Transformer and Mamba architectures, thereby forming a hybrid structure that simultaneously benefits from strong global modeling capacity and efficient sequence processing. Notably, the Transformer-Mamba framework has demonstrated promising results in the medical imaging domain, which inspires its adaptation to the depth completion task \cite{yanhetero}.


Specifically, the feature maps extracted from the RGB-D and depth branches, denoted as $\mathbf{F}_r$ and $\mathbf{F}_d$, are first fused using the previously described SMFM module:
\begin{equation}
\label{equa_smfm}
    \mathbf{F}_{f} = \text{SMFM}(\mathbf{F}_r, \mathbf{F}_d),
\end{equation}

Here, $\mathbf{F}_{f}$ represents the enhanced fusion feature, highlighting the most discriminative channels during cross-modal alignment.

To further capture long-range dependencies and model global contextual information, the fused feature is processed by the Transformer-Mamba block. The self-attention mechanism is defined as:
\begin{equation}
\label{equa2}
    \mathbf{F}_{f} = \text{LN}\big(\mathbf{F}_{f} + \text{MHA}(\mathbf{F}_{f})\big),
\end{equation}

Where $\text{MHA}(\cdot)$ denotes multi-head attention and $\text{LN}(\cdot)$ represents layer normalization.

Following self-attention, a state-space model (SSM) is employed to refine the fused features while capturing temporal or sequential dependencies. The hierarchical channel transformation within SSM is formulated as:
\begin{equation}
\label{equa3}
    \mathbf{F}_{f} = W_\text{down}\Big(\text{SSM}\big(\delta(\text{Conv}(W_\text{up}\mathbf{F}_{f}))\big)\Big) \odot  \delta(W_\text{up}\mathbf{F}_{f}),
\end{equation}

Where $W_\text{up}$ and $W_\text{down}$ denote channel expansion and compression projections, 
$\text{Conv}(\cdot)$ refers to a convolution layer, 
$\delta(\cdot)$ stands for the SiLU activation function, 
and $\odot$ denotes element-wise product.

Finally, a feed-forward network further enhances the feature representation:
\begin{equation}
\label{equa4}
    \mathbf{F}_{f} = \text{LN}\big(\mathbf{F}_{f} + \text{MLP}(\mathbf{F}_{f})\big),
\end{equation}

Where $\text{MLP}(\cdot)$ denotes a multi-layer perceptron. 

Through this module, the model 
can effectively integrate complementary information 
from the RGB-D and depth modalities, 
thereby producing fused features 
with enhanced representational capacity 
and enabling robust and precise depth completion.

\subsection{Decoder with Multi-Scale Fusion}


In depth completion tasks, 
effectively leveraging multi-scale features extracted 
from different stages of the encoder 
is crucial for improving prediction accuracy. 
Prior studies have demonstrated that progressively fusing cross-scale features can substantially improve the modeling capacity of the network\cite{fan2024tdcnet}. 
The decoder part of our network is largely consistent with \cite{fan2024tdcnet}, primarily integrating multi-scale feature information through a down-fusion module, 
which incorporates both channel attention and spatial attention mechanisms.

Given the feature maps from two adjacent layers, denoted as $\mathbf{F}_{i}$ and $\mathbf{F}_{i-1}$ with $i \in \{1,2,3\}$, we first downsample and channel-replicate the shallow feature $\mathbf{F}_{i-1}$ to match the size of $\mathbf{F}_{i}$:
\begin{equation}
\label{eq_fusion1}
    \hat{\mathbf{F}}_{i-1} = \text{CR}(\text{AAP}(\mathbf{F}_{i-1})),
\end{equation}

Where $\text{AAP}(\cdot)$ denotes adaptive average pooling and $\text{CR}(\cdot)$ represents channel repetition. The aligned features are then aggregated as

\begin{equation}
\label{eq_fusion2}
    \mathbf{F} = \mathbf{F}_{i} + \hat{\mathbf{F}}_{i-1}.
\end{equation}

To enhance discriminative information, we apply two complementary attention mechanisms. Spatial attention is defined as
\begin{equation}
\label{eq_fusion3}
    \mathbf{F}_{sa} = \text{Conv}_{7\times7}\big([\text{GMP}(\mathbf{F}), \text{GAP}(\mathbf{F})]\big),
\end{equation}

Where $\text{GAP}(\cdot)$ and $\text{GMP}(\cdot)$ denote global average pooling and global max pooling, respectively, and $\text{Conv}_{7\times7}$ is a convolution with a $7\times7$ kernel.  
Channel attention is formulated as:
\begin{equation}
\label{eq_fusion4}
    \mathbf{F}_{ca} = \text{Conv}_{1\times1}\big(\text{Conv}_{1\times1}(\text{GAP}(\mathbf{F}))\big).
\end{equation}

Next, the attended weights from spatial and channel domains are combined and applied to balance the contributions of different scales:

\begin{equation}
\label{eq_w}
    W = \sigma (\text{DWConv}([\mathbf{F}_{ca}+\mathbf{F}_{sa},F]))
\end{equation}

\begin{equation}
\label{eq_fusion5}
    \mathbf{F}^{\prime} = \text{Conv}_{1\times1}\big(W \otimes \mathbf{F}_{i}\big) 
    + \text{Conv}_{1\times1}\big((1-W) \otimes \hat{\mathbf{F}}_{i-1}\big)
\end{equation}

Where $\otimes$ denotes element-wise multiplication, $\text{DWConv}(\cdot)$ indicates depth-wise convolution and $W$ is the adaptive fusion weight.

The resulting feature is combined with 
$\mathbf{F}$ via an element-wise summation to form the residual connection representation:

\begin{equation}
\label{eq_fusion6}
    \mathbf{F}_{out} = \mathbf{F}^{\prime} + \mathbf{F}
\end{equation}

\subsection{Loss function}
Following prior work on depth completion \cite{fang2022transcg,fan2024tdcnet}, 
we adopt a training objective composed of two complementary loss terms. 
The first term is a pixel-wise regression loss, which computes the mean squared error (MSE) 
between the predicted depth map $D$ and the ground-truth depth map $D^*$, 
serving as the main supervisory signal:
\begin{equation}
\mathcal{L}_{\text{MSE}} = \frac{1}{N} \sum_{i=1}^{N} \left\| D_i - D^*_i \right\|^2,
\end{equation}

Where $N$ is the number of valid pixels.

In addition, to further enhance geometric consistency, 
we incorporate a normal-guided regularization term that encourages smoothness 
by aligning the predicted surface normals $\mathbf{n}(D)$ with the ground-truth normals $\mathbf{n}(D^*)$:
\begin{equation}
\mathcal{L}_{\text{normal}} = \frac{1}{N} \sum_{i=1}^{N} \left( 1 - \mathbf{n}(D_i) \cdot \mathbf{n}(D^*_i) \right),
\end{equation}

Where $\cdot$ denotes the dot product. 

The overall training objective is the weighted sum of the two terms:
\begin{equation}
\mathcal{L} = \mathcal{L}_{\text{MSE}} + \lambda \mathcal{L}_{\text{normal}},
\end{equation}

Where $\lambda$ is a hyperparameter balancing the contributions of the two losses.

\section{Experimental}
\subsection{Datasets and Metric}
In the training and evaluation of our model, we employ 2 publicly available datasets, namely TransCG\cite{fang2022transcg} and ClearGrasp\cite{sajjan2020clear}.  
The TransCG dataset is captured in real-world environments using a robotic platform and includes a total of 57,715 RGB-D images along with corresponding annotations, acquired from two different cameras. It encompasses 51 categories of everyday objects, many of which exhibit reflective, transparent, or translucent properties, and some possess complex hollow structures, posing significant challenges for depth sensing.
The ClearGrasp dataset is synthetically generated via the Synthesis AI platform using 9 transparent plastic CAD models. It provides 23,524 training images, while its test split includes both synthetic and real-world transparent objects, thereby enabling evaluation across both seen and unseen object shapes.  

To assess the effectiveness of our depth completion approach, 
we adopt four widely used measures: RMSE, REL, MAE, 
and threshold accuracy $\delta$. 
RMSE captures the root mean squared error between prediction and ground truth, 
while REL reflects their absolute relative error. 
MAE provides the overall mean absolute error:

\begin{equation}
\label{eq:rmse}
RMSE=\sqrt{\frac{1}{|\hat{D}|} \sum_{d \in \hat{D}} \| d - d^{*} \|^{2}}
\end{equation}

\begin{equation}
\label{eq:rel}
REL=\frac{1}{|\hat{D}|} \sum_{d \in \hat{D}} \frac{|d - d^{*}|}{d^{*}}
\end{equation}

\begin{equation}
\label{eq:mae}
MAE=\frac{1}{|\hat{D}|} \sum_{d \in \hat{D}} \lvert d - d^{*} \rvert
\end{equation}

Here, $d$ denotes the predicted depth, and $d^{*}$ represents the corresponding ground-truth depth.

Threshold accuracy $\delta$ denotes the percentage of pixels that satisfy $\max\!\left(\tfrac{d}{d^*}, \tfrac{d^*}{d}\right) < \delta$, where $d$ and $d^*$ are the predicted and reference depths. In our evaluation, $\delta$ values are set to 1.05, 1.10, and 1.25.

\subsection{Implementation Details}
In the experiments, model training and evaluation are performed using a single NVIDIA RTX 4090 GPU. 
Input images are resized to \(320 \times 240\) pixels.
The model is trained for 40 epochs using the AdamW optimizer with an initial learning rate of \(0.001\) and a batch size of 8. 
As illustrated in Fig. 2, the channel dimension 
$C$ is configured to 24.

\subsection{Dataset Results and Analysis}
\begin{figure*}[htbp] 
    \centering
    \includegraphics[width=\textwidth]{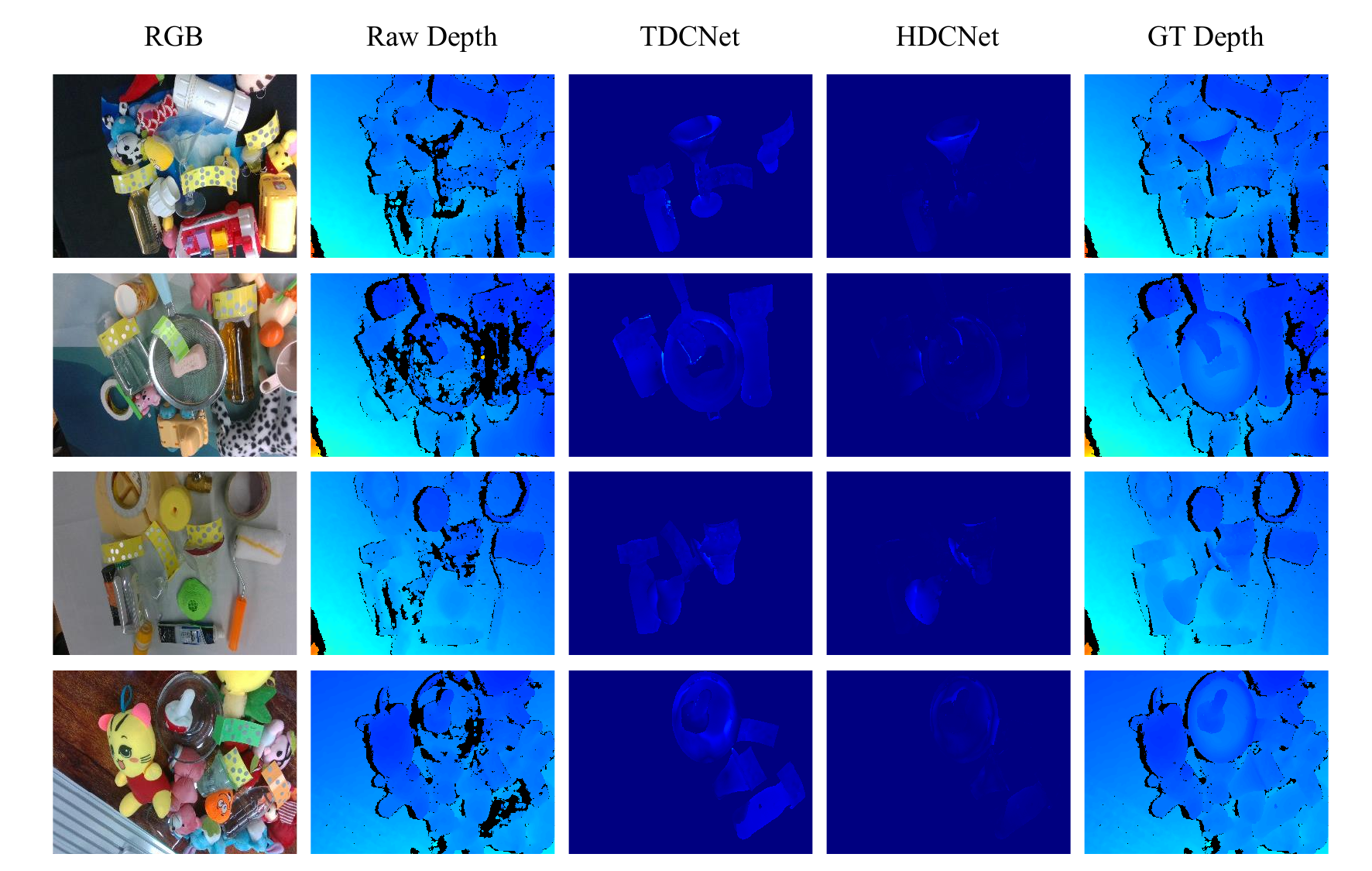} 
    \caption{Depth Completion Visualizations of Different Models on the TransCG Dataset}
    \label{fig_transcg}
\end{figure*}

\begin{figure*}[htbp] 
    \centering
    \includegraphics[width=\textwidth]{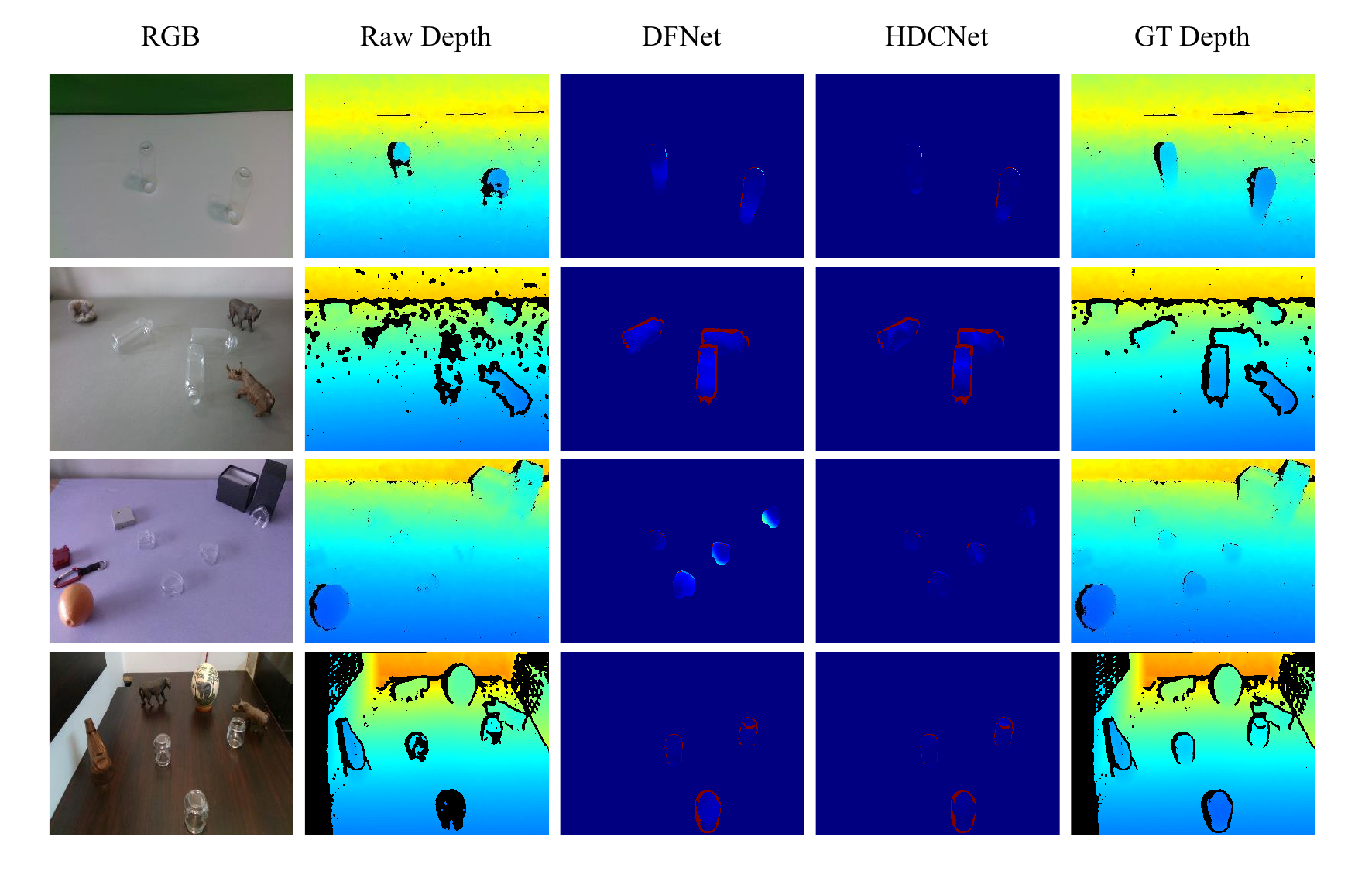} 
    \caption{Depth Completion Visualizations of Different Models on the ClearGrasp Real-world Dataset}
    \label{fig_cg}
\end{figure*}

In this section, we present the evaluation results of our model on the employed datasets and provide a systematic analysis. 
We also compare our approach with several baseline methods to highlight its advantages. 
Additionally, qualitative error visualizations 
are provided to demonstrate the model's performance 
on challenging transparent and reflective objects. 
Specifically, the evaluation results on the TransCG and ClearGrasp datasets are summarized in Tab.\ref{tab1} and Tab.\ref{tab2}, respectively. Experiments on these public datasets indicate that our proposed depth completion model achieves state-of-the-art performance. The corresponding error visualizations for each dataset are shown in Fig.\ref{fig_transcg} and Fig.\ref{fig_cg}, further confirming the superiority of our network. Compared with other baseline models, our approach exhibits lower errors and stronger depth completion capabilities.

The evaluation results on various public depth completion datasets demonstrate that our proposed model achieves state-of-the-art performance and exhibits strong generalization capability. Moreover, the visualization results reveal that, compared with baseline models, our approach yields smaller depth completion errors. Taken together, both the quantitative evaluations and qualitative visualizations consistently validate the superiority of our model.

\begin{table}[htbp]
    \centering
    \caption{Performance comparison of different methods on TransCG dataset}
    \label{tab1}
    \resizebox{\columnwidth}{!}{
    \begin{tabular}{lcccccc}
    \toprule
    \textbf{Methods} & \textbf{RMSE} $\downarrow$ & \textbf{REL} $\downarrow$ & \textbf{MAE} $\downarrow$ & \textbf{$\delta_{1.05}$}$\uparrow$ & \textbf{$\delta_{1.10}$}$\uparrow$ & \textbf{$\delta_{1.25}$}$\uparrow$ \\
    \midrule
    CG\cite{sajjan2020clear}         & 0.054  & 0.083  & 0.037  & 50.48  & 68.68  & 95.28  \\
    DFNet\cite{fang2022transcg}      & 0.018  & 0.027  & 0.012  & 83.76  & 95.67  & 99.71  \\
    LIDF\cite{zhu2021rgb} & 0.019  & 0.034  & 0.015  & 78.22  & 94.26  & 99.80  \\
    TCRNet\cite{zhai2024tcrnet}     & 0.017  & 0.020  & 0.010  & 88.96  & 96.94  & $\mathbf{99.87}$  \\
    TranspareNet\cite{xu2021seeing} & 0.026  & 0.023  & 0.013  & 88.45  & 96.25  & 99.42  \\
    FDCT\cite{li2023fdct}       & 0.015  & 0.022  & 0.010  & 88.18  & 97.15  & 99.81  \\
    TODE-Trans\cite{chen2023tode}  & $0.013$  & 0.019  & $\mathbf{0.008}$  & 90.43  & 97.39  & 99.81  \\
    DualTransNet\cite{liu2024transparent} & 0.012  & 0.018  & $\mathbf{0.008}$  & $92.37$  & 97.98  & 99.81  \\
    TDCNet\cite{fan2024tdcnet} &$\mathbf{0.012}$ &$\mathbf{0.017}$ &$\mathbf{0.008}$ &92.25 &97.86 &99.84 \\
    \midrule
    HDCNet (ours) & $\mathbf{0.012}$ & $\mathbf{0.017}$ & $\mathbf{0.008}$ & $\mathbf{92.70}$ & $\mathbf{98.09}$ & $\mathbf{99.89}$ \\
    \bottomrule
    \end{tabular}}
\end{table}

\begin{table}[ht]
        \centering
        \caption{Comparison of Different Methods on Real-world ClearGrasp Dataset.}
        \resizebox{\columnwidth}{!}{
        \begin{tabular}{lcccccc}
        \toprule
        \textbf{Method} & \textbf{RMSE} & \textbf{REL} & \textbf{MAE} & $\boldsymbol{\delta_{1.05}}$ & $\boldsymbol{\delta_{1.10}}$ & $\boldsymbol{\delta_{1.25}}$ \\
        \midrule
        CG~\cite{sajjan2020clear} & 0.037 & 0.049 & 0.027 & 74.30 & 88.47 & 96.27 \\
        
        DFNet~\cite{fang2022transcg}         & 0.026 & 0.037 & 0.020 & 76.69 & 92.26 & 99.09 \\
        FDCT\cite{li2023fdct}       & 0.028  & 0.038  & 0.021  & 76.45  & 93.36  & 98.95  \\
        TDCNet\cite{fan2024tdcnet} &$0.022$ &$0.031$ &$0.017$ &82.26 &95.83 &$\mathbf{99.85}$ \\
        \midrule
        HDCNet(ours) & $\mathbf{0.021}$ & $\mathbf{0.028}$ & $\mathbf{0.016}$ & $\mathbf{84.60}$ & $\mathbf{96.21}$ & $99.72$ \\
        \bottomrule
        \end{tabular}}
        \label{tab2}
\end{table}


\subsection{Robotic Grasping Experiments}

To validate the practical effectiveness 
and engineering applicability 
of the proposed depth completion network 
in robotic grasping tasks, systematic grasping experiments 
are conducted on transparent and reflective objects. 
In these experiments, the depth completion network functions as a depth perception module placed before the two-finger grasp detection network\cite{fang2023anygrasp}, providing more complete and reliable scene depth information to support high-precision grasp pose estimation. The experimental objects include 9 common transparent and reflective items, which are randomly arranged on a tabletop. For each object, the robotic arm performs five consecutive grasping attempts to compute the grasp success rate and evaluate the stability of the model performance.

The experimental setup employs a Franka Emika Panda seven-degree-of-freedom robotic arm equipped with its native two-finger gripper. The two-finger grasp detection network is built upon the AnyGrasp framework. Since the AnyGrasp framework was originally developed as an algorithm verification platform for different types of two-finger grippers, its default grasp pose coordinate system differs slightly from that of the native Franka gripper. To account for hardware discrepancies, minor translational and rotational adjustments are applied to the gripper coordinate system output by AnyGrasp during system integration, ensuring geometric and kinematic consistency with the Franka gripper.

The experimental results 
are summarized in the Tab.\ref{table_grasp}. As observed, 
depth completion substantially 
improves the grasping success rate for transparent objects. 
For reflective objects, where depth information 
is partially missing, the original grasp detection network 
can already predict grasp points reasonably well; 
however, incorporating depth completion further 
enhances the success rate, demonstrating the effectiveness 
of our network in practical grasping scenarios.

\begin{figure}[htbp] 
    \centering
    \includegraphics[width=\columnwidth]{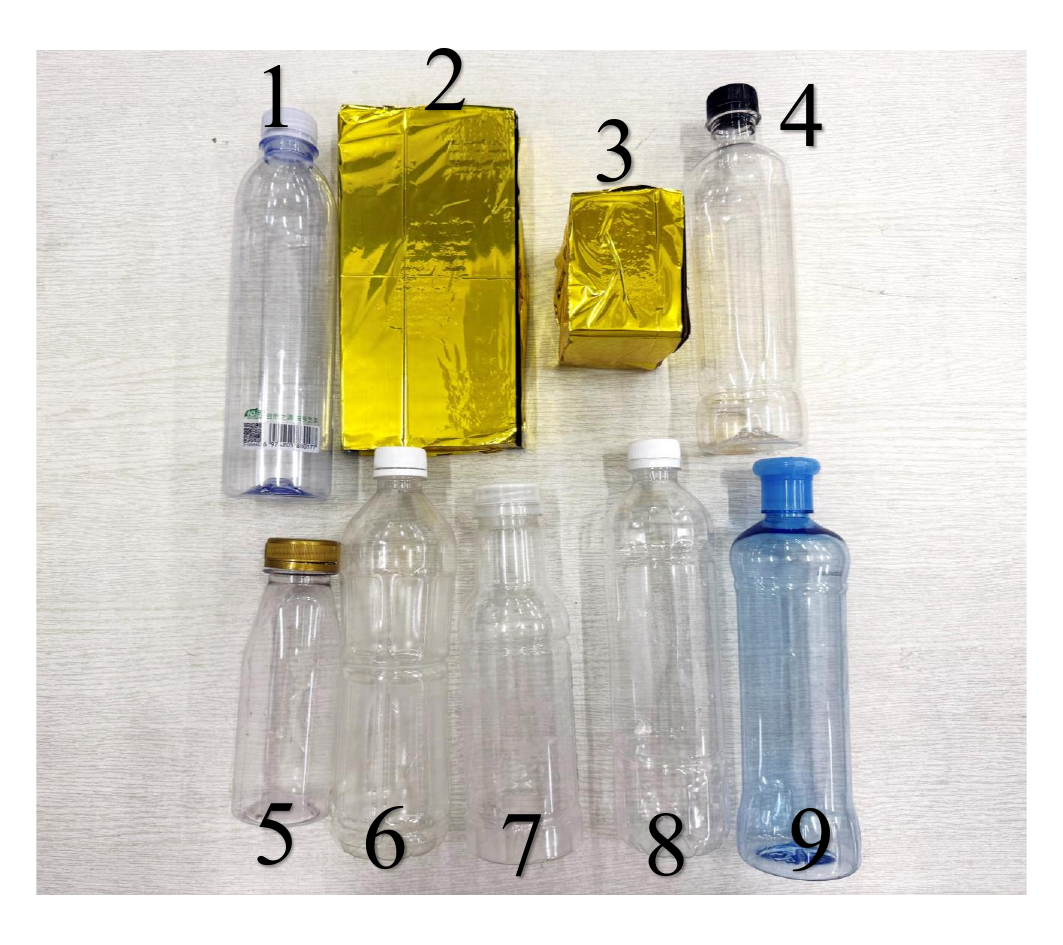} 
    \caption{Transparent and reflective objects in real-world grasping experiments.
    According to the sequence numbers, the objects are respectively: water bottle 1, reflective foam board, reflective box, beverage bottle 1, milk bottle, water bottle 2, beverage bottle 2, water bottle 3, and detergent bottle.}
    \label{grasp_objs}
\end{figure}

\begin{figure}[htbp] 
    \centering
    \includegraphics[width=\columnwidth]{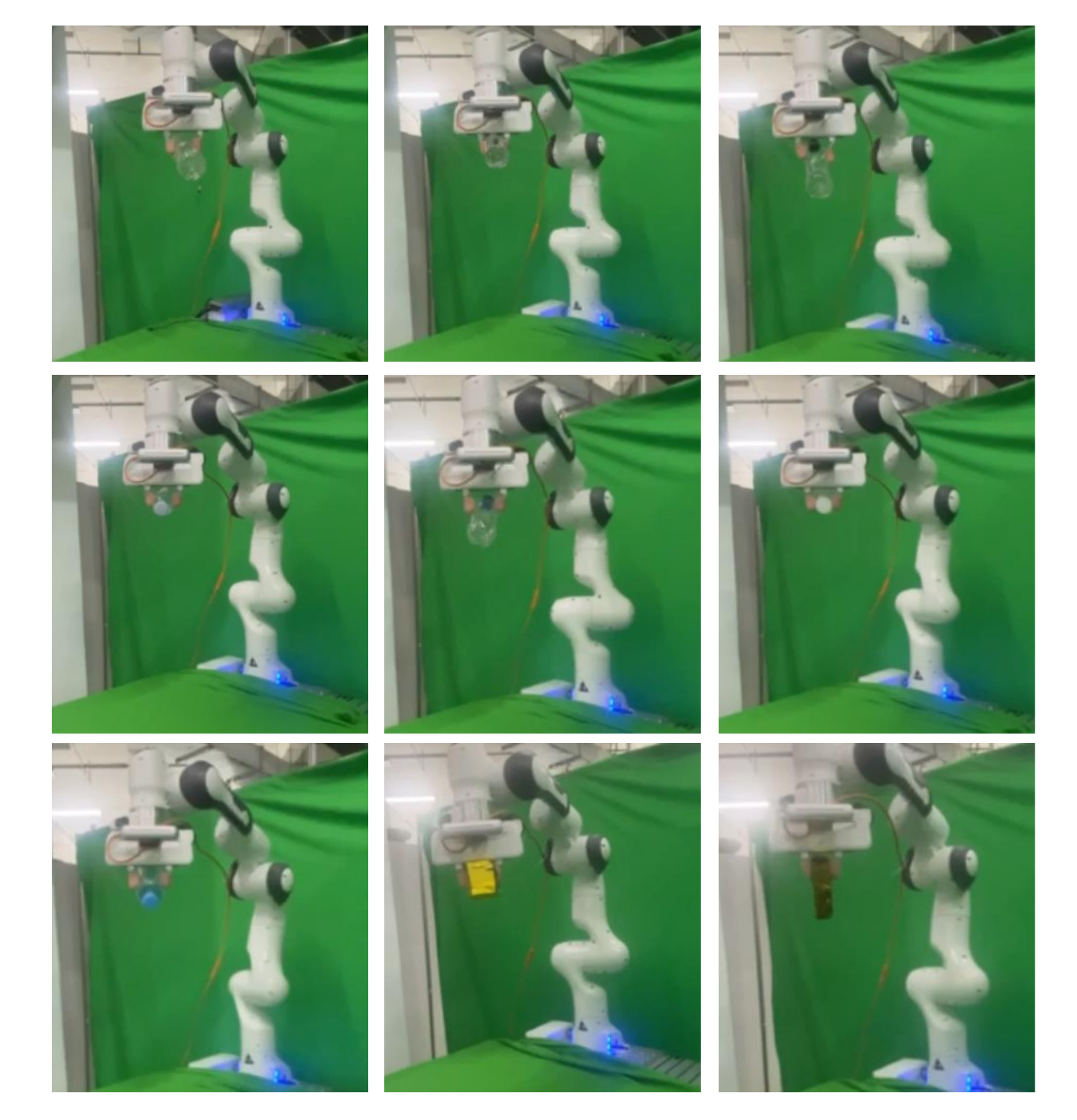} 
    \caption{Illustrative examples of grasping experiments.}
    \label{fig_realgrasp}
\end{figure}

\begin{table}[ht]
    \centering
    \caption{Grasping experiments in real-world scenes.}
    \resizebox{\columnwidth}{!}{
    \begin{tabular}{lcccc}
    \toprule
    
\textbf{Objs} & \textbf{AnyGrasp} & \textbf{HTMNet+AnyGrasp} \\
\midrule
1. water bottle 1     & 0/5 & 4/5 \\
2. reflective foam board          & 3/5 & 5/5 \\
3. reflective box  & 1/5 & 1/5 \\
4. beverage bottle 1             & 0/5 & 5/5 \\
5. milk bottle           & 0/5 & 2/5 \\
6. water bottle 2   & 1/5 & 4/5 \\
7. beverage bottle 2                & 1/5 & 4/5 \\
8. water bottle 3        & 1/5 & 5/5 \\
9. detergent bottle          & 0/5 & 4/5 \\
\midrule
\textbf{Success Rate}       & $15.6\%$ & \textbf{75.6\%} \\ \bottomrule

    \end{tabular}
    }
    \label{table_grasp}
\end{table}

\section{Ablation Study}
\subsection{Effectiveness Analysis of Multimodal Fusion Modules}
We conduct an ablation study on the proposed shallow and deep multimodal fusion modules to investigate their specific contributions to the depth completion task. As shown in Tab.\ref{tab5}, the experimental results demonstrate that incorporating either the shallow or the deep fusion module significantly improves the overall performance of the model. Specifically, the shallow fusion module facilitates the effective alignment of low-level geometric and textural features by fully exploiting early-stage multimodal information, while the deep fusion module enhances the integration of high-level semantic features across modalities, thereby improving the accuracy and robustness of depth estimation from a global perspective. These results collectively verify the effectiveness and complementarity of the proposed multimodal fusion mechanism at different feature levels.

\begin{table}[htbp]
    \centering
    \caption{The Role of Different Modules on TransCG Dataset}
    \label{tab5}
    \resizebox{\columnwidth}{!}{
    \begin{tabular}{cccccccc}
    \toprule
    \textbf{SMFM} &\textbf{BTMFM} & \textbf{RMSE} $\downarrow$ & \textbf{REL} $\downarrow$ & \textbf{MAE} $\downarrow$ & \textbf{$\delta_{1.05}$}$\uparrow$ & \textbf{$\delta_{1.10}$}$\uparrow$ & \textbf{$\delta_{1.25}$}$\uparrow$ \\
    \midrule
     & & $0.012$ & $0.019$ & $0.008$ & $92.38$ & $98.19$ & $99.89$ \\
    $\surd$  & & $0.012$ & $0.018$ & $0.008$ & $92.52$ & $\mathbf{98.18}$ & $99.89$ \\   
      &$\surd$ & $0.012$ & $0.019$ & $0.008$ & $92.21$ & $97.99$ & $99.89$ \\
    $\surd$  &$\surd$ & $\mathbf{0.012}$ & $\mathbf{0.017}$ & $\mathbf{0.008}$ & $\mathbf{92.70}$ & $98.09$ & $\mathbf{99.89}$ \\
    \bottomrule
    \end{tabular}}
\end{table}

\section{Conclusion}
In this work, we present HDCNet, a hybrid depth completion network designed to address the long-standing challenge of perceiving transparent and reflective objects in robotic manipulation. By leveraging the complementary advantages of Transformer, CNN, and Mamba architectures, HDCNet effectively bridges local structural cues and global contextual dependencies through a hierarchical multimodal fusion strategy. The proposed shallow fusion module facilitates the integration of fine-grained geometric and appearance information, while the deep Transformer–Mamba hybrid module enables robust semantic and contextual understanding.
Extensive experiments on multiple benchmark datasets verify that HDCNet consistently delivers state-of-the-art performance in depth completion, achieving both higher accuracy and stronger generalization capability. Moreover, real-world robotic grasping experiments demonstrate that the enhanced depth perception provided by HDCNet translates directly into improved manipulation performance, yielding a significant increase in grasp success rates for challenging transparent and reflective objects.
Overall, this study not only advances the reliability of depth perception under complex visual conditions but also highlights the potential of hybrid multimodal fusion architectures for future robotic perception and manipulation systems.

\bibliographystyle{IEEEtran}
\bibliography{refs}

\end{document}